\documentclass[11pt]{article}

\usepackage{acl}

\usepackage{tikz}
\usetikzlibrary{arrows.meta, positioning, shapes.geometric, fit, backgrounds}

\usepackage{times}
\usepackage{latexsym}
\usepackage[T1]{fontenc}
\usepackage[utf8]{inputenc}
\usepackage{microtype}
\usepackage{inconsolata}
\usepackage{graphicx}
\usepackage{amsmath}
\usepackage{amssymb}
\usepackage{booktabs}
\usepackage{multirow}
\usepackage{placeins}
\usepackage{float}
\usepackage{url}
\usepackage{xspace}
\usepackage{seqsplit}
\DeclareRobustCommand{\code}[1]{\texttt{\seqsplit{#1}}}

\usepackage{listings}
\usepackage{needspace}
\newcommand{\codeblockspace}{\Needspace{12\baselineskip}}
\lstdefinestyle{appendixcode}{
  basicstyle=\ttfamily\tiny,
  breaklines=true,
  breakatwhitespace=false,
  columns=fullflexible,
  keepspaces=true,
  frame=single,
  xleftmargin=0pt,
  xrightmargin=0pt,
  linewidth=\columnwidth,
  aboveskip=0.5em,
  belowskip=0.5em
}

\title{Guarded Repair for Harm-Aware Post-hoc Replacement of LLM Mathematical Reasoning}

\author{
Haizhou Xia \\
\texttt{haizhoux@outlook.com}
}

\newcommand{\method}{\textsc{GuardedRepair}\xspace}
\newcommand{\gsm}{GSM8K\xspace}
\newcommand{\asdiv}{ASDiv\xspace}
\newcommand{\dsflash}{\code{deepseek-v4-flash}\xspace}
\newcommand{\dspro}{\code{deepseek-v4-pro}\xspace}
\newcommand{\qwenone}{\code{qwen2.5:1.5b}\xspace}
\newcommand{\qwenseven}{\code{qwen2.5:7b}\xspace}
\newcommand{\qwenfourteen}{\code{qwen2.5:14b}\xspace}
\newcommand{\sgraph}{surface semantic-risk graph\xspace}
\newcommand{\repository}{\url{https://github.com/Haizhoux0517/guarded-repair}}

\begin{document}
\maketitle

\begin{center}
\small Code and artifacts: \repository
\end{center}

\begin{abstract}
Post-hoc repair of LLM mathematical reasoning introduces an asymmetric risk: fixing an incorrect reasoning trace is useful, but replacing a trace that was already correct can be harmful. We study this problem under a selective replacement setting, where a system must decide whether a repaired candidate is safer than preserving the original cached trace.

We present \method, a guarded best-of-$N$ repair framework that diagnoses cached reasoning traces, selectively triggers repair, and accepts answer-changing candidates only when deterministic verification guards support replacement. The framework combines lightweight symbolic checks, surface semantic-risk diagnostics, bounded candidate generation, and conservative acceptance policies.

On the full \gsm test set, where the initial reasoner already achieves 95.60\% accuracy, \method improves final accuracy to 96.89\%, fixing 17 of 58 remaining errors without measured broken-correct cases in the main run. On a weak-reasoner \asdiv setting, accuracy improves from 78.40\% to 87.60\%. Direct regeneration baselines show that this gain is not explained by stronger-model re-solving alone: re-solving all \gsm examples lowers accuracy to 93.03\% and breaks 47 initially correct answers. Additional analyses show that guarded repair substantially improves the fixed/broken tradeoff, while also revealing that replacement risk is reduced rather than eliminated.

These results support viewing post-hoc repair as harm-aware selective replacement rather than unconstrained re-solving.
\end{abstract}

\section{Introduction}

Large language models (LLMs) are strong mathematical word-problem solvers, but their reasoning traces can still contain arithmetic slips, missing constraints, semantic misbindings, or malformed final answers. These errors are difficult to handle by final-answer evaluation alone: a trace can be fluent while solving the wrong problem, and a repair model can confidently replace an initially correct answer with an incorrect one.

We study \emph{post-hoc repair}: an initial LLM has already produced a cached reasoning trace and final answer, and the system must decide whether to preserve or replace that trace. This setting differs from ordinary re-solving, self-correction, or best-of-$N$ selection because the original trace may already be correct. The central risk is therefore not only whether a new candidate is plausible, but whether accepting it is safer than keeping the original. We formulate this as harm-aware selective replacement: maximize fixed errors while controlling broken-correct cases.

\method instantiates this view as a guarded best-of-$N$ replacement protocol. The individual ingredients arithmetic checking, surface semantic heuristics, multiple candidates, and deterministic filters are not claimed to be new on their own. The contribution is to make the replacement decision explicit. The trigger controls when additional compute is spent and how many repair opportunities are exposed; the candidate generator provides a bounded set of possible replacements; the acceptance guards decide whether an answer-changing candidate is safer than the cached trace; and the evaluation protocol reports both fixed and broken transitions. This decomposition matters because a system can improve final accuracy while still causing unacceptable replacement harm. The goal is therefore not to introduce a new verifier, but to study how post-hoc repair should be accepted and evaluated when the original trace may already be correct.

On full \gsm, where the initial \dsflash reasoner leaves only 58 errors, \method fixes 17 of them and improves accuracy from 95.60\% to 96.89\% in the main run. On a seed-42 numeric \asdiv subset with \qwenone as the initial reasoner, it improves accuracy from 78.40\% to 87.60\%, fixing 92 errors. Additional \asdiv seeds, \gsm repair-stage reruns, weak-reasoner checks on SVAMP and MultiArith, and local Qwen repair-model checks show consistent positive net gains, while also revealing rare broken-correct cases and lower repair recall for weaker repair models. These results support the intended claim: guarded repair substantially reduces replacement risk relative to direct regeneration, but does not eliminate that risk universally and remains sensitive to candidate quality.

Our contributions are:
\begin{itemize}
    \item We identify replacement risk as a central failure mode in post-hoc mathematical-reasoning repair: a repair system must be evaluated not only by what it fixes, but also by what it breaks.
    \item We formalize repair as harm-aware selective replacement over cached traces, separating the trigger, candidate generator, acceptance guard, and fixed/broken evaluation roles.
    \item We instantiate this formulation with a guarded best-of-$N$ protocol and evaluate it with fixed/broken accounting, accepted-repair precision, candidate-flow analysis, and compute-aware direct-regeneration baselines.
    \item We show on \gsm, numeric \asdiv, and weak-reasoner robustness settings that guarded repair improves the fixed/broken tradeoff over direct regeneration, while clarifying the remaining limits of repair safety and candidate quality.
\end{itemize}
\section{Related Work}

\paragraph{Reasoning and verification.}
Chain-of-thought and decomposition prompts improve mathematical reasoning by encouraging intermediate steps \citep{wei2022chain,zhou2023least}. Verifier and process-supervision methods score candidate solutions or intermediate steps \citep{cobbe2021training,lightman2023verify}; recent process-reward work extends this with learned process-level evaluators \citep{she2025rprm}. \method shares the goal of checking reasoning, but performs post-hoc replacement of an existing cached trace and uses deterministic diagnostics rather than a learned verifier.

\paragraph{Self-correction, test-time scaling, and tools.}
Self-refinement and self-correction ask models to critique or revise their own outputs \citep{madaan2023selfrefine,wu2025stepco,xiong2025selfrewarding,zhang2025selfverify}. Test-time scaling and best-of-$N$ methods spend additional inference to sample, rank, or revise solutions. Tool-augmented methods such as ReAct, PAL, and program-of-thought prompting offload computation to external actions or executable programs \citep{yao2023react,gao2023pal,chen2023program}. In contrast, \method does not re-solve every problem or convert it into a program; it repairs cached natural-language traces only when replacement appears safer than preservation.

\paragraph{Selective prediction and math benchmarks.}
Selective prediction studies when a system should abstain under uncertainty \citep{elyaniv2010selective,geifman2017selective}. Our action is different: the system already has an answer and must decide whether to replace it. Prior math word-problem work emphasizes the need to model quantities and relations \citep{hosseini2014learning,patel2021nlp}; \asdiv provides a diverse arithmetic corpus \citep{miao2020diverse}. We evaluate on full \gsm and numeric \asdiv, and Appendix~\ref{app:related-positioning} summarizes how \method differs from adjacent paradigms. Adjacent paradigms mainly decide which newly generated solution to trust, how to revise a solution through model feedback, or when to abstain. \method instead starts from an already cached trace and asks a different question: whether an answer-changing repair is safer than preserving the original trace. This makes broken-correct accounting central rather than auxiliary.
\section{Method}

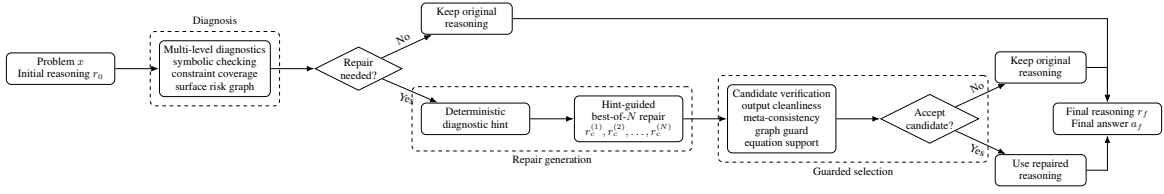
\begin{figure*}[t]
\centering
\resizebox{0.96\textwidth}{!}{
\begin{tikzpicture}[
    node distance=1.15cm and 1.25cm,
    box/.style={
        rectangle,
        rounded corners,
        draw=black,
        thick,
        align=center,
        minimum width=3.1cm,
        minimum height=0.9cm,
        font=\small
    },
    smallbox/.style={
        rectangle,
        rounded corners,
        draw=black,
        thick,
        align=center,
        minimum width=2.6cm,
        minimum height=0.75cm,
        font=\small
    },
    decision/.style={
        diamond,
        draw=black,
        thick,
        align=center,
        aspect=2.0,
        inner sep=1.5pt,
        font=\small
    },
    arrow/.style={-{Latex[length=2.2mm]}, thick},
    dashedarrow/.style={-{Latex[length=2.2mm]}, thick, dashed},
    group/.style={
        rectangle,
        rounded corners,
        draw=black,
        dashed,
        inner sep=0.25cm
    }
]

\node[box] (input) {Problem $x$\\Initial reasoning $r_0$};
\node[box, right=of input] (diag) {Multi-level diagnostics\\
\footnotesize symbolic checking\\
\footnotesize constraint coverage\\
\footnotesize surface risk graph};

\node[decision, right=of diag] (trigger) {Repair\\needed?};

\node[smallbox, above right=0.65cm and 1.15cm of trigger] (keep1) {Keep original\\reasoning};

\node[box, below right=0.65cm and 1.15cm of trigger] (hint) {Deterministic\\diagnostic hint};

\node[box, right=of hint] (bestofn) {Hint-guided\\best-of-$N$ repair\\
\footnotesize $r_c^{(1)}, r_c^{(2)}, \ldots, r_c^{(N)}$};

\node[box, right=of bestofn] (verify) {Candidate verification\\
\footnotesize output cleanliness\\
\footnotesize meta-consistency\\
\footnotesize graph guard\\
\footnotesize equation support};

\node[decision, right=of verify] (accept) {Accept\\candidate?};

\node[smallbox, above right=0.65cm and 1.15cm of accept] (keep2) {Keep original\\reasoning};

\node[smallbox, below right=0.65cm and 1.15cm of accept] (repair) {Use repaired\\reasoning};

\node[box, right=2.1cm of accept] (output) {Final reasoning $r_f$\\Final answer $a_f$};

\draw[arrow] (input) -- (diag);
\draw[arrow] (diag) -- (trigger);

\draw[arrow] (trigger) -- node[above, sloped, font=\footnotesize] {No} (keep1);
\draw[arrow] (trigger) -- node[below, sloped, font=\footnotesize] {Yes} (hint);

\draw[arrow] (hint) -- (bestofn);
\draw[arrow] (bestofn) -- (verify);
\draw[arrow] (verify) -- (accept);

\draw[arrow] (accept) -- node[above, sloped, font=\footnotesize] {No} (keep2);
\draw[arrow] (accept) -- node[below, sloped, font=\footnotesize] {Yes} (repair);

\draw[arrow] (keep1.east) -| (output.north);
\draw[arrow] (keep2.east) -| (output.north);
\draw[arrow] (repair.east) -| (output.south);

\begin{scope}[on background layer]
\node[group, fit=(diag), label={[font=\footnotesize]above:Diagnosis}] {};
\node[group, fit=(hint)(bestofn), label={[font=\footnotesize]below:Repair generation}] {};
\node[group, fit=(verify)(accept), label={[font=\footnotesize]below:Guarded selection}] {};
\end{scope}

\end{tikzpicture}
}
\caption{Guarded best-of-$N$ post-hoc repair. The default action is to keep the cached trace; replacement occurs only when a triggered repair candidate passes deterministic guards.}
\label{fig:pipeline}
\end{figure*}

\subsection{Problem Formulation}

Given a problem $x$, cached initial reasoning $r_0$, and answer $a_0$, the system outputs final reasoning $r_f$ and answer $a_f$. Gold answers are used only for evaluation. Let $F$ be the number of initially wrong examples fixed by repair and $B$ the number of initially correct examples broken by repair. We view post-hoc repair as constrained replacement:
\begin{equation}
    \max_{\pi} F(\pi) \quad \text{s.t.} \quad B(\pi) \leq \epsilon.
\end{equation}
The default action is to keep the original trace. A replacement is made only when a candidate passes the guarded acceptance policy:
\begin{equation}
 r_f =
 \begin{cases}
 r_c, & \text{if a candidate repair passes all gates},\\
 r_0, & \text{otherwise}.
 \end{cases}
\end{equation}

This framing decomposes repair into distinct decision components. The trigger $T(x,r_0)$ determines whether additional compute is spent and therefore controls candidate recall and cost. The generator $G_N(x,r_0)$ constructs a bounded candidate set of size at most $N$. The acceptance rule $A(x,r_0,r_c)$ implements the harm constraint by allowing answer-changing replacements only when deterministic evidence supports the candidate over the cached trace. Finally, the evaluation protocol reports both $F$ and $B$, because higher final accuracy can be misleading if it is obtained by breaking many initially correct traces.

\subsection{Diagnostics and Triggering}

\method is a replacement-risk decision protocol rather than a standalone verifier, parser, or decoding algorithm. It instantiates the decomposition above with lightweight deterministic diagnostics, bounded best-of-$N$ candidate generation, and conservative acceptance guards. The guards are intentionally simple and auditable: their role is not to solve semantic parsing or verification as standalone tasks, but to operationalize the replacement decision under an explicit harm constraint. The diagnostics include arithmetic equation checking, numeric constraint coverage, meta-consistency scoring, and \sgraph signals. The \sgraph is a conservative risk feature rather than a semantic parser: it extracts quantity mentions and surface relation patterns such as aggregation, comparison, rate, change-event, and part-whole relations, then emits risk categories such as quantity binding, comparison warnings, answer-format warnings, or per-entity rate omissions. Details, thresholds, risk categories, and a manual audit of graph-risk signals are in Appendices~\ref{app:answer-norm}, \ref{app:semantic-graph-algorithm}, \ref{app:semantic-taxonomy}, \ref{app:semantic-graph-human-audit}, and~\ref{app:thresholds}.

Repair is attempted only when diagnostics indicate risk, including empty generations, arithmetic failures, high-risk semantic issues, severe missing-constraint signals, or low meta-consistency. Otherwise, the cached trace is preserved without calling the repair model.

\subsection{Best-of-$N$ Repair and Guarded Acceptance}

For each triggered case, the repair model generates up to $N$ JSON-formatted candidates using different prompt strategies: hint-guided repair, strict concise arithmetic, and solving from the original problem while treating the initial trace as a warning signal. The final experiments use $N=3$. Each candidate is normalized, re-diagnosed, and evaluated by deterministic gates for output cleanliness, semantic-graph risk, meta-consistency, and equation support. The policy does not use a learned verifier or LLM judge. In relaxed-support settings, the final answer must be explicitly supported by an arithmetic or number-theoretic derivation. Pseudocode, prompts, cleanliness checks, equation-support grammar, and ablation switches are in Appendices~\ref{app:bestofn}, \ref{app:prompts}, \ref{app:cleanliness}, \ref{app:eq-support}, \ref{app:accept-policy}, and~\ref{app:ablation-configs}.
\section{Experimental Setup}

\paragraph{Datasets and models.}
The primary evaluation uses the full \gsm test split with 1,319 samples. The weak-reasoner evaluation uses a uniformly sampled 1,000-example numeric \asdiv subset drawn with seed 42 from a 2,147-example numeric pool. We sample to control repair-stage cost; the subset is drawn before any repair experiment and is not selected by model performance. To assess sampling sensitivity, Appendix~\ref{app:asdiv-seeds} reports three additional 1,000-example seeds from the same pool. The filtering and sampling procedure is given in Appendix~\ref{app:asdiv-filter}. \gsm initial traces are generated by \dsflash, while the main repair candidates use \dspro. In weak-reasoner settings, the initial reasoner is \qwenone. We also run local \qwenseven and \qwenfourteen repair-model portability checks on the \asdiv seed-42 setting while keeping the cached initial traces and acceptance policy fixed. Repair calls use temperature 0.0, JSON output, a 768-token budget, and one 512-token format retry. Model and decoding details are in Appendix~\ref{app:model-details}.

\paragraph{Metrics.}
We report initial/final accuracy, absolute improvement, fixed errors, broken-correct cases, accepted-repair precision, error repair rate, and harm rate:
\begin{equation}
    \mathrm{HarmRate}=\frac{\#\mathrm{BrokenCorrect}}{\#\mathrm{Total}}.
\end{equation}
Accepted repair precision is the fraction of accepted repairs that are wrong-to-correct transitions. This is conservative: accepted wrong-to-wrong modifications lower precision even if they do not break correct answers. We also report exact paired sign-test probabilities for changed examples and use ``zero measured harm'' only descriptively; with zero observed broken-correct cases, the rule-of-three gives approximate 95\% upper bounds of 0.23\% for \gsm and 0.30\% for the 1,000-sample \asdiv setting.
\section{Results}

\subsection{Main Results}

Table~\ref{tab:main-results} shows the main results. On full \gsm, \method improves accuracy from 95.60\% to 96.89\%, fixing 17 of 58 initially incorrect cases with zero measured broken-correct cases in the main run. Although the absolute \gsm gain is 1.29 points, this setting starts from a high baseline; the result corresponds to correcting 29.3\% of the remaining initial errors without measured harm in the main run. On the weak-reasoner \asdiv setting, it improves accuracy from 78.40\% to 87.60\%, fixing 92 errors with zero measured broken-correct cases. The paired sign-test probabilities for changed examples are $1.53\times 10^{-5}$ on \gsm and $4.04\times 10^{-28}$ on \asdiv.

\begin{table*}[t]
\centering
\small
\begin{tabular}{llrrrrrr}
\toprule
Dataset & Initial reasoner & Initial Acc. & Final Acc. & $\Delta$ & Fixed & Broken & Harm \\
\midrule
\gsm full test & \dsflash & 95.60 & 96.89 & +1.29 & 17 & 0 & 0.00 \\
\asdiv numeric-1000 & \qwenone & 78.40 & 87.60 & +9.20 & 92 & 0 & 0.00 \\
\bottomrule
\end{tabular}
\caption{Main results. Accuracy and improvement are reported in percentage points. In the high-baseline \gsm setting, \method fixes 17 of 58 remaining errors; in the weak-reasoner \asdiv setting, it fixes 92 errors. Both main runs introduce zero measured broken-correct cases. Under the rule of three, zero observed harm implies approximate 95\% upper bounds of 0.23\% on \gsm and 0.30\% on \asdiv numeric-1000.}
\label{tab:main-results}
\end{table*}

The larger \asdiv gain reflects greater repairable error mass: \gsm has 58 initially wrong cases, while the weak \asdiv setting has 216. The \asdiv result is not unique to seed 42: across four uniformly sampled 1,000-example seeds, the mean improvement is +9.05 points, with gains ranging from +8.50 to +9.80 points. We report seed-level mean, standard deviation, and range rather than treating the seed-42 subset as a full-pool estimate. Additional checks in the appendix support the robustness and boundaries of this result: \gsm repair-stage reruns remain above the initial accuracy but show rare broken-correct cases (Appendix~\ref{app:stability-reruns}); additional \asdiv seeds also reveal rare harm (Appendix~\ref{app:asdiv-seeds}); and SVAMP/MultiArith weak-reasoner checks improve final accuracy while each introducing one broken-correct case (Appendix~\ref{app:weak-robustness}). Repair behavior and accepted-outcome accounting are in Appendices~\ref{app:repair-behavior-details} and~\ref{app:accepted-outcomes}.

\subsection{Ablations}

Table~\ref{tab:gsm-ablation} reports \gsm ablations. Increasing from $N=1$ to $N=3$ improves repair recall, but fixed/broken accounting is non-monotonic: $N=2$ fixes more than $N=1$ but introduces two broken-correct cases. Removing the semantic graph gate reduces final accuracy and fixes fewer errors, while disabling equation-support verification introduces two broken-correct cases and lowers accuracy. Relaxed missing-constraint acceptance reaches the same final accuracy as the main system but accepts more repairs without fixing more errors, so we keep the stricter policy. These results form a discrete fix--harm frontier: the selected configuration is not the one that accepts the most repairs, but the one with the best observed fixed/broken tradeoff.

\begin{table*}[t]
\centering
\small
\begin{tabular}{lrrrrrr}
\toprule
Setting & Final Acc. & $\Delta$ & Fixed & Broken & Accepted & Attempts \\
\midrule
Initial only & 95.60 & -- & 0 & 0 & 0 & 0 \\
Guarded $N=1$ & 96.06 & +0.45 & 6 & 0 & 9 & 508 \\
Guarded $N=2$ & 96.13 & +0.53 & 9 & 2 & 13 & 1007 \\
Guarded $N=3$ & 96.89 & +1.29 & 17 & 0 & 20 & 1498 \\
$N=3$ without graph gate & 96.66 & +1.06 & 14 & 0 & 16 & 1501 \\
$N=3$ without equation support & 96.59 & +0.99 & 15 & 2 & 22 & 1487 \\
$N=3$ with relaxed missing-constraint acceptance & 96.89 & +1.29 & 17 & 0 & 22 & 1494 \\
\bottomrule
\end{tabular}
\caption{Ablation results on the full \gsm test split. Accuracy and improvement are reported in percentage points. The final system uses guarded $N=3$ repair with surface graph diagnostics and equation-support verification.}
\label{tab:gsm-ablation}
\end{table*}

\subsection{Direct Strong-Model Baselines}

Table~\ref{tab:direct-baselines} tests whether gains come merely from invoking the stronger repair model. Directly re-solving all \gsm examples with the strong model lowers accuracy to 93.03\% and breaks 47 initially correct solutions. Re-solving only triggered examples reduces cost but still breaks 7 correct cases. The closest compute-controlled baseline, direct best-of-3 solving with the same triggered set, strong model, number of calls, and gates but without the initial trace or diagnostic hint, reaches 96.29\%, fixes 12 errors, and breaks 3 correct cases. Under the same 1,498-call budget, \method reaches 96.89\%, fixes 17 errors, and introduces zero measured broken-correct cases.

\begin{table*}[t]
\centering
\small
\begin{tabular}{lrrrrr}
\toprule
Method & Strong calls & Relative calls & Final Acc. & Fixed & Broken \\
\midrule
Initial only & 0 & 0.00x & 95.60 & 0 & 0 \\
Strong solve all, $N=1$ & 1319 & 1.00x & 93.03 & 13 & 47 \\
Strong solve triggered, $N=1$ & 508 & 0.39x & 96.06 & 13 & 7 \\
Direct best-of-3 + gates & 1498 & 1.14x & 96.29 & 12 & 3 \\
\method, $N=3$ & 1498 & 1.14x & 96.89 & 17 & 0 \\
\bottomrule
\end{tabular}
\caption{Compute-aware strong-model baselines on the full \gsm test split. Relative calls are normalized by one full direct solve pass over all 1,319 examples. Direct strong-model regeneration either hurts accuracy or introduces broken-correct cases. Under the same 1,498-call budget as direct best-of-3 with gates, \method fixes more errors and introduces no measured broken-correct cases.}
\label{tab:direct-baselines}
\end{table*}
\subsection{Open-Source Repair-Model Portability}

To test whether the guarded replacement protocol depends entirely on the proprietary API repair model, we replace the repair model in the \asdiv seed-42 setting with local Qwen2.5 repair models while keeping the same cached \qwenone initial traces and deterministic acceptance policy. Table~\ref{tab:open-repair-portability} shows that both local repair models preserve positive gains without measured broken-correct cases, but their gains are smaller than \dspro. This suggests that the protocol transfers beyond a single API repair model, while repair recall remains strongly dependent on candidate quality.

\begin{table}[t]
\centering
\scriptsize
\setlength{\tabcolsep}{2.4pt}
\begin{tabular}{@{}lrrrrr@{}}
\toprule
Repair & Final & $\Delta$ & Fix & Brk. & Acc. \\
\midrule
DeepSeek & 87.60 & +9.20 & 92 & 0 & 95 \\
Qwen-7B & 81.10 & +2.70 & 27 & 0 & 36 \\
Qwen-14B & 80.70 & +2.30 & 23 & 0 & 30 \\
\bottomrule
\end{tabular}
\caption{Open-source repair-model portability check on the \asdiv seed-42 setting. DeepSeek denotes \dspro; Qwen-7B and Qwen-14B denote local \qwenseven and \qwenfourteen. The initial reasoner is fixed to \qwenone, cached initial traces are unchanged, and only the repair model is replaced. Local Qwen repair models yield smaller but positive gains without measured broken-correct cases.}
\label{tab:open-repair-portability}
\end{table}

\section{Analysis}

\paragraph{Why guarding is necessary.}
The ablations show that more candidate generation is not enough: $N=2$ increases fixes but introduces broken-correct cases, and disabling equation support accepts more repairs while lowering final accuracy. A typical unsafe candidate copies or derives a plausible number without explicitly supporting the final answer, or omits a late constraint while producing a clean-looking derivation. The guarded policy therefore requires a candidate to be safer than preserving the initial trace, not merely plausible.

\paragraph{Candidate flow and remaining errors.}
Candidate-flow logs separate generation failures from conservative rejections. On \gsm, 25 initially wrong examples obtain at least one correct candidate and 17 are accepted; on \asdiv, 130 obtain a correct candidate and 92 are accepted. The remaining errors come from both candidate-generation limits and guarded-acceptance false rejections. This explains the main tradeoff: stricter gates reduce harm but reject some correct candidates. Appendix~\ref{app:candidate-flow-details} reports the full flow table, and Appendix~\ref{app:qualitative-examples} provides representative fixed-error, rejected-unsafe-repair, and false-rejection cases.

\paragraph{Cost.}
Best-of-3 uses 1,498 repair attempts on \gsm and 1,859 on \asdiv because repair is triggered only for suspicious traces rather than for every example. This corresponds to 1.14 strong-model calls per \gsm example and 1.86 calls per \asdiv example. The trigger rate is 38.5\% on \gsm (508/1,319) and 67.3\% on \asdiv (673/1,000), while the accepted-repair rate among triggered examples is 3.94\% and 14.1\%, respectively. Appendix~\ref{app:cost-practicality} discusses how $N$ controls the recall--cost tradeoff.
\section{Limitations and Threats to Validity}

The study is limited to arithmetic word problems with mostly numeric final answers; geometry, algebra proofs, program synthesis, and open-ended reasoning may require different normalization and diagnostics. The \sgraph checker is heuristic rather than a complete parser: it is designed as a conservative risk-control feature for quantity binding, comparison, answer-format, and related risks, not as a high-precision semantic-understanding module. It can miss subtle errors and frequently emits benign warnings. We therefore evaluate it through downstream ablations, candidate-flow analysis, qualitative examples, diagnostic logs, and a small stratified manual audit rather than as a standalone parser.

Most non-main configurations use a single run, and API-based repair generation may vary even at temperature 0.0. For the main \gsm setting, Appendix~\ref{app:stability-reruns} reports three additional repair-stage reruns with fixed cached initial traces; all remain positive relative to the 95.60\% initial accuracy, but rare broken-correct cases appear. Thus zero measured harm in the main \gsm and seed-42 \asdiv runs should not be interpreted as proof that the system cannot break correct answers. Additional SVAMP, MultiArith, \asdiv multi-seed, and \gsm rerun checks show that the current guards reduce but do not eliminate replacement risk. The local Qwen repair-model check further shows that the protocol is not tied to a single API repair model, but also that repair recall is model-dependent.

Finally, the main weak-reasoner \asdiv result uses a 1,000-example sample from a 2,147-example numeric pool to control repair-stage cost. We mitigate sampling concerns by reporting three additional uniformly sampled seeds, which show consistent positive gains, but this is still not a full-pool evaluation. The multi-seed results should be interpreted as a sampling-sensitivity check rather than a replacement for evaluating all 2,147 numeric examples. The numeric filtering also excludes yes/no, comparison, and other non-numeric answer types, leaving broader answer formats for future work.
\section{Reproducibility Statement}

We release evaluation code, answer-normalization scripts, diagnostic rules, guarded acceptance policy, prompts, thresholds, dataset split files, cached initial traces, repair candidate logs, diagnostic outputs, and final prediction files at \repository. These artifacts allow repair-stage analysis to be reproduced without regenerating initial model outputs. For \asdiv numeric-1000, we will release full numeric-pool IDs, sampled IDs for all reported seeds, filtering scripts, sampling seeds, and rejected-category counts, so that the pre-run uniform sampling procedure can be audited. We also release the 50-case stratified semantic-graph audit file, its manual annotations, and the corresponding summary statistics under \code{artifacts/semantic\_graph\_audit/gsm8k\_main/}. We also report local Qwen repair-model runs to provide a reproducible non-API portability check. API model names, local model names, decoding parameters, token budgets, JSON schema, and retry policy are reported in Appendices~\ref{app:model-details}, \ref{app:prompts}, and~\ref{app:thresholds}.
\section{Conclusion}

We presented \method as a harm-aware selective replacement protocol for post-hoc repair of LLM mathematical reasoning traces. The system diagnoses cached traces, repairs only risky cases, and accepts replacements only when deterministic guards support changing the original answer. Across \gsm, numeric \asdiv, and weak-reasoner robustness checks, the method improves final accuracy and substantially reduces harm relative to direct strong-model regeneration. Multi-seed, rerun, robustness, and local repair-model experiments also clarify the boundary: guarded repair reduces replacement risk, but rare broken-correct cases can still occur and repair recall depends on candidate quality. These findings support treating post-hoc repair as selective replacement: fixing wrong traces matters, but a repair system should also be judged by whether it preserves traces that were already correct.

\clearpage
\appendix
\raggedright

\section{Additional Results and Positioning}
\label{app:additional-results}

\subsection{Positioning Relative to Prior Paradigms}
\label{app:related-positioning}
Table~\ref{tab:related-positioning} summarizes the main distinction from adjacent paradigms. Unlike methods that re-solve, rerank, or abstain, \method decides whether replacing an existing cached trace is safer than preserving it.

\begin{table}[H]
\centering
\scriptsize
\setlength{\tabcolsep}{3pt}
\begin{tabular}{@{}p{0.31\columnwidth}p{0.61\columnwidth}@{}}
\toprule
Prior paradigm & Difference from this work \\
\midrule
Verifier reranking & Optimizes candidate correctness, not replacement harm against an existing trace. \\
Self-consistency & Re-solves the problem and does not preserve a cached correct trace by default. \\
Self-correction & May replace the original answer without explicit fixed/broken accounting. \\
Process reward models & Require trained reward or verifier models; our guards are deterministic diagnostics. \\
Tool/program reasoning & Converts the task into executable reasoning rather than repairing cached natural-language traces. \\
Selective prediction & Abstains under uncertainty; our setting decides whether replacement is safer than preservation. \\
\bottomrule
\end{tabular}
\caption{Positioning of \method relative to related paradigms. The key distinction is harm-aware selective replacement over an existing cached trace.}
\label{tab:related-positioning}
\end{table}

\subsection{Repair Behavior Details}
\label{app:repair-behavior-details}
This section gives the full repair-behavior accounting referenced in Section~\ref{tab:main-results}. The table reports how often repair is triggered, how many candidate replacements are accepted, and how many accepted repairs are true wrong-to-correct fixes.

\begin{table}[H]
\centering
\scriptsize
\setlength{\tabcolsep}{3pt}
\begin{tabular}{@{}lrrrrrr@{}}
\toprule
Dataset & Trig. & Acc. & Fix & Brk. & Att. & Prec. \\
\midrule
\gsm & 508 & 20 & 17 & 0 & 1498 & 85.00 \\
\asdiv & 673 & 95 & 92 & 0 & 1859 & 96.84 \\
\bottomrule
\end{tabular}
\caption{Repair behavior under guarded best-of-3. Accepted precision is the percentage of accepted repairs that are wrong-to-correct transitions; accepted repairs that remain wrong lower precision even when they do not create harm.}
\label{tab:repair-behavior}
\end{table}
\FloatBarrier

\subsection{Weak-Reasoner Robustness Results}
\label{app:weak-robustness}
This section expands the weak-reasoner robustness summary from the main text. All rows use \qwenone as the initial reasoner and the same guarded best-of-3 repair stage.

\begin{table}[H]
\centering
\scriptsize
\setlength{\tabcolsep}{3pt}
\begin{tabular}{@{}lrrrrrr@{}}
\toprule
Dataset & Init. & Final & $\Delta$ & Fix & Brk. & Harm \\
\midrule
\asdiv & 78.40 & 87.60 & +9.20 & 92 & 0 & 0.00 \\
SVAMP & 73.67 & 88.00 & +14.33 & 44 & 1 & 0.33 \\
MultiArith & 96.67 & 98.33 & +1.67 & 4 & 1 & 0.56 \\
\bottomrule
\end{tabular}
\caption{Weak-reasoner robustness checks. The method improves all three weak-initial settings, with larger gains when more repairable error mass is available. Harm is the percentage of all examples that become incorrect after initially being correct.}
\label{tab:weak-robustness}
\end{table}
\FloatBarrier

\subsection{Candidate Flow Details}
\label{app:candidate-flow-details}
This section decomposes the repair pipeline over initially wrong examples. The counts separate errors that remain because no correct candidate is generated from errors that remain because a correct candidate is rejected by the guarded acceptance policy.

\begin{table}[H]
\centering
\tiny
\setlength{\tabcolsep}{2pt}
\begin{tabular}{@{}lrrrrrrrr@{}}
\toprule
Data & InitW & TrigW & CorrC & AccC & RejC & NoC & FinalW & Brk \\
\midrule
\gsm & 58 & 43 & 25 & 17 & 8 & 33 & 41 & 0 \\
\asdiv & 216 & 166 & 130 & 92 & 38 & 86 & 124 & 0 \\
\bottomrule
\end{tabular}
\caption{Candidate flow over initially wrong examples. InitW/TrigW are initially wrong and triggered-wrong counts; CorrC/AccC/RejC are correct-candidate, accepted-correct, and rejected-correct counts.}
\label{tab:candidate-flow}
\end{table}
\FloatBarrier

\subsection{Qualitative Repair Examples and Outcome Taxonomy}
\label{app:qualitative-examples}
Table~\ref{tab:case-study} gives representative \gsm repair outcomes selected from the analysis logs. The cases illustrate that the guarded policy is not only a final-answer filter. It can accept a repair when the candidate restores a missing semantic relation, reject a plausible but unsafe answer change when the original is already correct, reject a correct candidate when residual diagnostic warnings remain, or accept a locally clean candidate that is still wrong. The last two cases correspond to the recall--precision boundary in Table~\ref{tab:candidate-flow}: stricter gates reduce broken-correct cases, but accepted precision is not perfect.

\begin{table*}[t]
\centering
\scriptsize
\begin{tabular}{@{}p{0.16\textwidth}p{0.20\textwidth}p{0.36\textwidth}p{0.20\textwidth}@{}}
\toprule
Case type & Initial issue & Candidate behavior & Gate decision \\
\midrule
Fixed error (\gsm id 186) & Initial answer binds the total straw quantity to the asked rat count. & Candidate separates hamster straw, rabbit straw, and rat straw, then solves \code{18r = 90} to obtain \code{5}. & Accepted: graph-clean repair with supported final answer. \\
Rejected unsafe repair (\gsm id 380) & Initial answer \code{803} is correct, but diagnostics trigger a repair attempt. & Candidate changes the answer to \code{760.42} by using a different interest interpretation and leaves residual graph/format warnings. & Rejected by residual graph/format risk: original answer preserved, avoiding a broken-correct case. \\
False rejection (\gsm id 510) & Initial answer \code{31} is wrong for the required daily calorie deficit. & Candidate computes \code{30 * 3500 = 105000} and \code{105000 / 200 = 525}, matching the gold answer. & Rejected: residual quantity-binding warning causes the system to keep the original. \\
Accepted but still wrong (\gsm id 340) & Initial answer \code{2} is wrong after a juggling setup with growth and dropped balls. & Candidate correctly grows from \code{3} to \code{7} balls, but treats two balls caught by the crowd as still available and subtracts only the permanently lost ball, yielding \code{6} instead of the gold \code{4}. & Accepted: locally clean and equation-supported, but still wrong; this lowers accepted precision without creating broken-correct harm. \\
\bottomrule
\end{tabular}
\caption{Representative qualitative repair outcomes. The examples show how guarded acceptance fixes some errors, avoids unsafe replacements, rejects some correct candidates, and occasionally accepts a still-wrong candidate when local evidence is insufficient.}
\label{tab:case-study}
\end{table*}

\subsection{Qualitative Outcome Taxonomy}
\label{sec:case-taxonomy}

Table~\ref{tab:case-taxonomy} summarizes the main qualitative outcome classes observed in the analysis logs. The table is not a separate human annotation study; rather, it aggregates the same fixed/broken accounting, candidate-flow logs, diagnostic categories, and rejection counters used in the quantitative analysis. It complements the individual examples in Table~\ref{tab:case-study} by showing where the repair pipeline succeeds and where it remains conservative.

\begin{table*}[t]
\centering
\scriptsize
\begin{tabular}{@{}p{0.17\textwidth}p{0.20\textwidth}rp{0.43\textwidth}@{}}
\toprule
Setting & Outcome class & Count & Diagnostic pattern \\
\midrule
\gsm & Accepted fixed repairs & 17 & Wrong-to-correct accepted repairs. The fixed set includes arithmetic/final-answer corrections as well as semantic-risk cases flagged by quantity binding, comparison, per-entity rate, or change-event diagnostics. \\
\gsm & Accepted but unsuccessful repairs & 3 & Wrong-to-wrong accepted repairs. These candidates changed the answer but did not reach the gold answer, lowering accepted precision without creating broken-correct harm. \\
\gsm & Correct candidates rejected & 8 & Initially wrong examples where at least one generated candidate had the gold answer but guarded acceptance rejected it, reflecting the recall cost of conservative gating. \\
\asdiv & Accepted fixed repairs & 92 & Weak-reasoner errors corrected mostly under low-symbolic-coverage, arithmetic-error, or quantity-binding diagnostics. \\
\asdiv & Correct candidates rejected & 38 & Correct candidates rejected under residual diagnostic risk, graph-guard rejection, formatting failure, or strict acceptance rejection. \\
SVAMP/MultiArith & Broken-correct robustness cases & 2 & Rare weak-reasoner over-repair cases: one SVAMP and one MultiArith initially correct answer is replaced by a plausible but incorrect candidate. \\
Direct best-of-3 baseline & Broken-correct direct regenerations & 3 & Same-budget direct solving with gates but without the initial trace or diagnostic hint breaks three initially correct \gsm examples, showing why direct regeneration is not sufficient. \\
\bottomrule
\end{tabular}
\caption{Qualitative outcome taxonomy from the analysis logs. Counts summarize where the repair system fixes errors, where it accepts still-wrong repairs, where conservative gates reject correct candidates, and where residual harm appears in robustness or direct-regeneration baselines.}
\label{tab:case-taxonomy}
\end{table*}

The taxonomy highlights three recurring patterns. First, the main gains come from answer-changing repairs that convert initially wrong examples into correct ones, especially in the weak-reasoner settings. Second, some generated correct candidates remain rejected; this is the main source of lost recall and explains why the system does not fix every initially wrong example. Third, the rare broken-correct cases in SVAMP, MultiArith, and the direct baseline arise from plausible answer-changing candidates that pass local checks but reinterpret a quantity relation differently from the gold solution. These cases support the paper's main design choice: repair should be treated as guarded selective replacement rather than unconstrained regeneration.

\subsection{Cost and Practicality}
\label{app:cost-practicality}
Best-of-3 increases inference cost, but the trigger policy avoids re-solving every example. On \gsm, repair is triggered for 508 of 1,319 examples (38.5\%) and makes 1,498 repair attempts, or 1.14 strong-model calls per dataset example. Only 20 triggered examples are accepted (3.94\% of triggered cases), reflecting the high baseline and conservative acceptance policy. On \asdiv, repair is triggered for 673 of 1,000 examples (67.3\%) and makes 1,859 attempts, or 1.86 calls per example; 95 triggered examples are accepted (14.1\%), reflecting the larger repairable error mass in the weak-reasoner setting.

The parameter $N$ controls a recall--cost tradeoff. Lower $N$ reduces strong-model calls but can miss correct repairs, while higher $N$ increases candidate recall and also increases the number of candidates that must be rejected by the guards. In applications where safety and auditability are more important than raw throughput, the selected $N=3$ setting is a conservative operating point; for lower-cost deployment, $N$ or the trigger threshold can be reduced.

\section{Implementation Details and Reproducibility}
\label{app:details}

This appendix summarizes the implementation details needed to reproduce the reported experiments. The main experiments use cached initial reasoning traces, so the repair stage can be rerun without regenerating the initial model outputs.

\subsection{Model and Decoding Configuration}
\label{app:model-details}

Table~\ref{tab:model-reporting} reports the model configuration used in the main experiments. The exact API snapshot may affect reproducibility, so the implementation caches initial reasoning traces and reports repair-stage decoding settings explicitly. For API-based models, provider-side snapshots may change over time; the reported results therefore rely on cached initial traces and saved repair candidate outputs. Re-running the repair-stage analysis from these cached artifacts does not require regenerating the initial model outputs.

\begin{table}[t]
\centering
\scriptsize
\setlength{\tabcolsep}{3pt}
\begin{tabular}{@{}lll@{}}
\toprule
Role & Model & Setting \\
\midrule
\gsm initial & DS-Flash & API, $T=0$ \\
Main repair & DS-Pro & API, $T=0$, 768 tok. \\
Local repair & Qwen-7B/14B & Ollama, $T=0$, 768 tok. \\
Weak initial & Qwen-1.5B & \asdiv/SVAMP/MultiArith, $T=0$ \\
\bottomrule
\end{tabular}
\caption{Model configuration. DS-Flash and DS-Pro denote \dsflash and \dspro; Qwen-1.5B, Qwen-7B, and Qwen-14B denote \qwenone, \qwenseven, and \qwenfourteen. The main repair model is used for the primary \gsm and \asdiv results; local Qwen models are used only for the repair-model portability check.}
\label{tab:model-reporting}
\end{table}

\subsection{Answer Extraction and Normalization}
\label{app:answer-norm}

The final-answer parser first searches for explicit answer markers such as \code{Final Answer:}, \code{final answer is}, and \code{answer:}. If such markers are found, it extracts the last answer-like value in the marked span. Supported answer forms include integers, decimals, comma-formatted numbers, fractions, colon-formatted time or ratio strings, and yes/no answers. If no explicit marker is found, the parser falls back to the last answer-like value in the reasoning trace, but the repair pipeline itself requires clean repair candidates to contain exactly one \code{Final Answer} line.

For evaluation, answers are normalized by stripping punctuation and unit parentheses, removing thousands separators, normalizing decimal integers such as \code{12.0} to \code{12}, and comparing exact normalized strings before applying numeric, fraction, and ratio equivalence. This normalization supports examples such as \code{1,059,955} versus \code{1059955}, \code{12} versus \code{12.0}, and \code{2/3} versus \code{4/6}.

\subsection{ASDiv Numeric Filtering}
\label{app:asdiv-filter}

The \asdiv numeric subset is constructed from a locally stored \asdiv-derived JSONL pool of 2,305 examples, converted into the common input format used by the pipeline. The preprocessing script first normalizes answers while preserving fractions, comma-formatted numbers, and colon-formatted time or ratio strings. We then keep only examples whose normalized answers are plain numeric values: integers, decimals, or fractions. We exclude yes/no answers, likely categorical question types, and non-numeric or ambiguous answer formats. Colon-formatted answers such as \code{2:3} or \code{12:50} are excluded because they may represent either ratios or times. This yields 2,147 numeric examples, from which we uniformly sample 1,000 examples with seed 42 before running any repair experiment. The 1,000-example subset is used to control repair-stage API cost and is not selected by model performance. To make the sampling decision auditable, we release the full numeric-pool IDs, sampled IDs for all reported seeds, filtering scripts, sampling seeds, and rejected-category counts.

\begin{table}[t]
\centering
\small
\begin{tabular}{lr}
\toprule
Filtering step & Examples \\
\midrule
Original local \asdiv-derived pool & 2305 \\
Rejected: non-numeric question type & 95 \\
Rejected: non-numeric/ambiguous answer format & 58 \\
Rejected: yes/no answer & 5 \\
Final numeric pool & 2147 \\
Sampled with seed 42 & 1000 \\
\bottomrule
\end{tabular}
\caption{Construction of the numeric \asdiv subset. The final 1,000-example evaluation set is uniformly sampled from the numeric pool with seed 42 before repair runs.}
\label{tab:asdiv-filtering}
\end{table}

\subsection{Surface Semantic-Risk Graph Algorithm}
\label{app:semantic-graph-algorithm}

The \sgraph checker is implemented with deterministic surface-pattern rules rather than an LLM, a learned verifier, or a dependency parser. The checker first extracts numeric mentions from the problem and reasoning trace after normalizing number words, decimals, fractions, commas, and simple money expressions. For each numeric mention, it stores a local token window and derives a lightweight node annotation consisting of the surface number, normalized value, unit phrase, entity mention, and local predicate context.

Unit phrases are extracted from nearby content words following or preceding the number, after removing stopwords and arithmetic function words. Entity mentions are approximated by nearby noun-like tokens and named entities in the same local window. Predicate context is represented by nearby verbs and comparative markers such as \code{more}, \code{fewer}, \code{left}, \code{total}, \code{each}, \code{per}, \code{gave}, \code{bought}, \code{removed}, and \code{remaining}. These annotations are diagnostic features rather than a full semantic parse.

Edges are added by deterministic relation templates. Aggregation edges are triggered by markers such as \code{total}, \code{together}, \code{in all}, or additive equations. Comparison edges are triggered by \code{more than}, \code{fewer than}, \code{less than}, and related comparative forms. Rate edges are triggered by \code{each}, \code{per}, \code{every}, and multiplicative contexts. Change-event edges are triggered by verbs such as \code{gave}, \code{lost}, \code{spent}, \code{removed}, \code{bought}, \code{received}, or \code{added}. Part-whole edges are triggered when a total quantity is paired with subgroup quantities.

The graph score is a heuristic risk score used only for triggering and guarded acceptance. Starting from 1.0, the checker subtracts penalties for high-risk mismatches, including quantity-binding conflicts, reversed or ignored comparison relations, missing rate multiplication, change-event direction errors, answer-format mismatches, and generation failures. The final score is clipped to $[0,1]$. The graph guard accepts a candidate only if it avoids generation failure, has no high-risk semantic issue, and its score is at least the configured minimum threshold. In the main experiments, this threshold is 0.60. This value is a conservative engineering threshold and is not tuned on test labels; we evaluate its downstream role through the semantic-graph ablation and the manual risk-signal audit.

\codeblockspace
\begin{lstlisting}[style=appendixcode]
def semantic_graph_check(problem, trace):
    q_problem = extract_quantities(problem)
    q_trace = extract_quantities(trace)

    g_problem = build_relation_graph(q_problem, problem)
    g_trace = build_relation_graph(q_trace, trace)

    risks = []
    risks += check_quantity_binding(g_problem, g_trace)
    risks += check_comparisons(g_problem, g_trace)
    risks += check_rate_usage(g_problem, g_trace)
    risks += check_change_events(g_problem, g_trace)
    risks += check_answer_format(problem, trace)

    score = 1.0
    for risk in risks:
        score -= penalty(risk)
    score = clip(score, 0.0, 1.0)

    return score, risks
\end{lstlisting}

\subsection{Surface Semantic-Risk Taxonomy}
\label{app:semantic-taxonomy}

Table~\ref{tab:semantic-taxonomy} summarizes the main \sgraph risk categories and their trigger conditions. The checker emits risk categories rather than correctness labels; these categories are used as trigger signals and as part of guarded candidate acceptance.

\begin{table*}[t]
\centering
\small
\begin{tabular}{lll}
\toprule
Risk type & Trigger condition & Example failure \\
\midrule
\code{quantity\_binding} & Same number used with different nearby entity/unit & Apples assigned to oranges \\
\code{comparison\_warning} & Comparative marker is reversed, ignored, or unsupported & Uses 5 instead of base $+5$ \\
\code{per\_entity\_rate\_missing} & \code{each}/\code{per} quantity lacks entity multiplier & $3$ bags $\times$ $4$ candies treated as $4$ \\
\code{change\_event\_misinterpretation} & Add/remove/spend/give event has wrong direction & Had 10, gave away 3 $\rightarrow$ uses 13 \\
\code{answer\_format\_warning} & Requested value type mismatches final answer type & Outputs total instead of difference \\
\bottomrule
\end{tabular}
\caption{Surface semantic-risk graph categories. The checker uses deterministic surface-pattern rules and emits heuristic risk signals rather than proof-level correctness labels.}
\label{tab:semantic-taxonomy}
\end{table*}

\subsection{Surface Semantic-Risk Graph Diagnostic Logs}
\label{app:semantic-graph-logs}

The \sgraph checker is not evaluated as a standalone semantic parser. Instead, we inspect the diagnostic-log patterns produced during the main \gsm run to understand how graph signals participate in triggering and acceptance decisions. These logs summarize acceptance-decision patterns rather than all individual model calls, so the counts below should be interpreted as implementation diagnostics, not as semantic-parser precision or recall.

\begin{table}[t]
\centering
\scriptsize
\begin{tabular}{lr}
\toprule
Diagnostic quantity & Count \\
\midrule
Acceptance-decision patterns inspected & 171 \\
Accepted patterns & 7 \\
Rejected patterns & 164 \\
No-op rejections & 156 \\
Answer-changing candidates & 15 \\
Answer-changing accepted & 7 \\
Answer-changing rejected & 8 \\
Accepted answer-changing candidates that are graph-clean & 7 \\
Candidate graph-clean patterns & 142 \\
Candidate graph-risk patterns & 29 \\
\bottomrule
\end{tabular}
\caption{Surface semantic-risk graph diagnostic-log summary on the \gsm main run. The table summarizes acceptance-decision patterns rather than a standalone semantic-parser evaluation. No-op candidates are rejected when the final answer does not change.}
\label{tab:semantic-graph-logs}
\end{table}

\begin{table}[t]
\centering
\scriptsize
\begin{tabular}{lr}
\toprule
Initial semantic graph risk type & Count \\
\midrule
\code{quantity\_binding\_error} & 106 \\
\code{quantity\_binding\_error+comparison\_warning} & 6 \\
\code{comparison\_warning} & 6 \\
\code{per\_entity\_rate\_missing} & 1 \\
\code{change\_event\_misinterpretation} & 1 \\
\code{answer\_format\_warning} & 1 \\
\code{none} & 50 \\
\bottomrule
\end{tabular}
\caption{Initial semantic graph risk types observed in the diagnostic-log patterns. The distribution shows that the graph checker primarily detects quantity-binding risks, while also covering comparison, rate, change-event, and answer-format warnings.}
\label{tab:semantic-risk-log-dist}
\end{table}

The log summary supports the interpretation of the \sgraph checker as a downstream risk-control module. In the inspected patterns, all accepted answer-changing candidates are graph-clean, while graph-risk patterns are dominated by quantity-binding errors and include comparison, rate, change-event, and answer-format warnings. This evidence complements the ablation in Table~\ref{tab:gsm-ablation}: removing the graph gate does not catastrophically increase measured harm in that run, but it reduces fixed errors and final accuracy. The rejected unsafe repair in Table~\ref{tab:case-study} illustrates the intended safety role: a candidate would change an initially correct answer, but residual graph/format warnings cause the system to preserve the cached trace.

\subsection{Manual Audit of Surface Graph Risk Signals}
\label{app:semantic-graph-human-audit}

To check whether \sgraph warnings correspond to human-identifiable risks, we manually audit 50 stratified non-none initial graph-risk cases from the main \gsm run. The sample is stratified over emitted risk types, so it is intended as a qualitative audit of risk coverage rather than an estimate of parser precision or end-to-end guard precision. The audit labels each case as a clear semantic risk, a conservative/benign but relevant warning, or no identifiable risk. Table~\ref{tab:semantic-graph-human-audit} summarizes the results.

\begin{table}[t]
\centering
\scriptsize
\begin{tabular}{lrr}
\toprule
Manual judgment & Count & Percent \\
\midrule
Clear semantic risk & 6 & 12.0 \\
Conservative/benign warning & 44 & 88.0 \\
No identifiable risk & 0 & 0.0 \\
\bottomrule
\end{tabular}
\caption{Manual audit of 50 stratified non-none surface graph risk cases from the main \gsm run. The audit checks whether emitted graph-risk signals correspond to identifiable semantic-risk patterns; it is not a standalone semantic-parser evaluation or a high-precision parser benchmark.}
\label{tab:semantic-graph-human-audit}
\end{table}

The audit covers quantity-binding warnings (25 cases), comparison warnings (12), combined quantity-binding/comparison warnings (7), answer-format warnings (3), and one case each of split interpretation, change-event interpretation, and per-entity rate omission. All audited warnings correspond to at least a conservative semantic-risk pattern, but most are benign warnings on traces that ultimately remain correct. This supports using the graph checker as an auditable risk-control signal while reinforcing that it should not be interpreted as a high-precision standalone parser. In particular, the 88\% conservative/benign-warning rate indicates that the graph feature is deliberately recall-oriented: it is useful for preventing unsafe replacement and shaping trigger decisions, but its warnings should be filtered by the downstream acceptance policy rather than treated as correctness judgments.

\subsection{Repair Trigger Policy}
\label{app:trigger-policy}

The medium-strength repair trigger is designed to avoid repairing every sample. In pseudocode, repair is attempted when any of the following conditions hold:
\begin{itemize}
    \item the initial reasoning is empty;
    \item the meta-diagnosis is \code{generation\_failure}, \code{arithmetic\_error}, or \code{logical\_contradiction};
    \item the semantic graph diagnosis is \code{generation\_failure};
    \item a high-risk semantic issue is detected, including \code{times\_more\_interpretation}, \code{per\_entity\_rate\_missing}, \code{equally\_split\_interpretation}, or \code{change\_event\_misinterpretation};
    \item the meta-diagnosis is \code{missing\_constraint} and the global meta score is below 0.90;
    \item the global meta score is below 0.65.
\end{itemize}
Otherwise, the original reasoning is preserved without calling the repair model.

\codeblockspace
\begin{lstlisting}[style=appendixcode]
def trigger(meta, graph, trace):
    if empty(trace): return True
    if meta.err in {gen_fail, arith_err, logic_err}:
        return True
    if graph.err == gen_fail:
        return True
    if high_risk_graph(graph):
        return True
    if meta.err == missing_constraint:
        return meta.score < 0.90
    return meta.score < 0.65
\end{lstlisting}

\subsection{Best-of-$N$ Repair Selection}
\label{app:bestofn}

For each triggered sample, the system generates up to $N$ repair candidates. The final experiments use $N=3$. Each candidate is converted from JSON into the standardized trace format and is then checked for output cleanliness, symbolic consistency, constraint coverage, meta-consistency, semantic graph quality, and equation support. The system selects the first candidate that passes both the semantic graph guard and the deterministic guarded acceptance policy. If no candidate is accepted, the final reasoning remains the original initial reasoning.

\codeblockspace
\begin{lstlisting}[style=appendixcode]
for i in range(N):
    cand = repair_llm(x, r0, diag0, i)
    if not clean(cand):
        continue
    diag_c = diagnose(x, cand)
    graph_ok = graph_guard(graph0, diag_c.graph)
    accept = accept_policy(r0, cand, diag0, diag_c)
    if graph_ok and accept:
        return cand
return r0
\end{lstlisting}

\subsection{Prompt Strategies}
\label{app:prompts}

The three repair attempts use the same JSON schema but different attempt strategies:
\begin{itemize}
    \item Attempt 0 uses the deterministic diagnostic hint as guidance when helpful and changes the answer only if the previous answer is not supported by the problem.
    \item Attempt 1 prioritizes strict formatting and concise arithmetic, and instructs the model to preserve the original answer if it is defensible.
    \item Attempt 2 solves from the original problem in at most four compact steps, treating the initial reasoning only as a warning signal.
\end{itemize}
All attempts require a JSON object with a \code{steps} list and a number-only \code{final\_answer}. The format retry uses the same schema and asks the model to rewrite only the malformed output, without prose outside JSON.

The repair prompt follows this skeleton:
\codeblockspace
\begin{lstlisting}[style=appendixcode]
System: Return only valid JSON. No markdown.

Schema:
{
  "steps": ["short arithmetic step"],
  "final_answer": "number"
}

Rules:
- Use at most 4 steps.
- Prefer arithmetic equations.
- final_answer is number-only.
- Use but do not mention the hint.
- Attempt style: <style>

Problem: <problem>
Initial: <initial reasoning>
Hint: <diagnostic hint>
Semantic error: <semantic error>
Meta error: <meta error>
\end{lstlisting}

\subsection{Candidate Cleanliness}
\label{app:cleanliness}

A candidate is considered clean only if it is non-empty, contains a parseable final answer, includes exactly one \code{Final Answer} line, and is not excessively long. The implementation also rejects candidates containing meta-discussion phrases such as \code{previous reasoning}, \code{the diagnosis says}, \code{provided hint}, \code{this is ambiguous}, or references to the prompt itself. This prevents verbose self-analysis from being accepted as a repaired mathematical solution.

\subsection{Equation-Support Guard}
\label{app:eq-support}

The equation-support guard checks whether the candidate final answer is explicitly supported by the candidate reasoning. It recognizes:
\begin{itemize}
    \item standard arithmetic equations, e.g., \code{276 / 12 = 23};
    \item LCM/GCD equations, e.g., \code{LCM(6, 5) = 30};
    \item restricted derivational statements such as \code{Number of trays = 23} or \code{The greatest common divisor is 15}.
\end{itemize}
The guard intentionally does not accept arbitrary copied quantities, such as \code{Time saved = 64}, because such statements may simply repeat a number from the problem without deriving the answer. The ablation without this guard introduces two broken-correct cases on \gsm, supporting its role as a safety condition.

\subsection{Accepted Repair Outcome Accounting}
\label{app:accepted-outcomes}

Accepted repair precision is defined as
\begin{equation}
    \frac{\#(\text{wrong}\rightarrow\text{correct accepted repairs})}
    {\#(\text{accepted repairs})}.
\end{equation}
This is a conservative measure: accepted repairs that do not fix an initially wrong example lower precision even if they do not create harm. Table~\ref{tab:accepted-outcomes} decomposes accepted repairs by before/after correctness transition.

\begin{table}[t]
\centering
\scriptsize
\setlength{\tabcolsep}{2.5pt}
\begin{tabular}{@{}lrrrrr@{}}
\toprule
Dataset & Acc. & W$\to$C & C$\to$W & W$\to$W & C$\to$C \\
\midrule
\gsm full test & 20 & 17 & 0 & 3 & 0 \\
\asdiv num.-1000 & 95 & 92 & 0 & 3 & 0 \\
\bottomrule
\end{tabular}
\caption{Accepted repair outcomes. W$\to$C repairs are fixed errors; C$\to$W repairs are broken-correct cases; W$\to$W repairs are accepted but unsuccessful modifications; C$\to$C repairs are unnecessary but non-harmful modifications.}
\label{tab:accepted-outcomes}
\end{table}

\subsection{ASDiv Multi-Seed Subset Check}
\label{app:asdiv-seeds}

The main \asdiv experiment uses a 1,000-example subset to control repair-stage cost. Because the filtered numeric pool contains 2,147 examples, sampling can introduce variance. We therefore run three additional uniformly sampled 1,000-example subsets using seeds 0, 1, and 2. These subsets are sampled from the same filtered numeric pool before repair is run, and no seed is selected based on repair performance. All runs use the same weak initial reasoner, repair model, and guarded best-of-3 configuration. Table~\ref{tab:asdiv-seeds} reports the original seed-42 run together with seeds 0, 1, and 2.

\begin{table}[t]
\centering
\scriptsize
\setlength{\tabcolsep}{2.5pt}
\begin{tabular}{lrrrrrr}
\toprule
Seed & Init. & Final & $\Delta$ & Fixed & Broken & Acc. \\
\midrule
42 & 78.40 & 87.60 & +9.20 & 92 & 0 & 95 \\
0 & 79.50 & 88.00 & +8.50 & 87 & 2 & 91 \\
1 & 79.90 & 88.60 & +8.70 & 91 & 4 & 97 \\
2 & 78.80 & 88.60 & +9.80 & 104 & 6 & 113 \\
\midrule
Mean & 79.15 & 88.20 & +9.05 & 93.50 & 3.00 & 99.00 \\
Std. & 0.59 & 0.42 & 0.50 & 6.34 & 2.24 & 8.37 \\
\bottomrule
\end{tabular}
\caption{ASDiv multi-seed subset check. Each row uses a uniformly sampled 1,000-example subset from the same 2,147-example numeric pool. All seeds show consistent positive gains, while additional seeds reveal rare broken-correct cases. ``Acc.'' denotes accepted repairs.}
\label{tab:asdiv-seeds}
\end{table}

Across four seeds, final accuracy ranges from 87.60\% to 88.60\%, and absolute improvement ranges from +8.50 to +9.80 percentage points, with mean improvement +9.05 and standard deviation 0.50. Thus, the \asdiv gain is not unique to seed 42. However, the additional seeds also introduce rare broken-correct cases, so the seed-42 zero-harm observation should not be interpreted as a universal property of the weak-reasoner setting. Since only four sampled subsets are evaluated and the subsets may overlap, we treat the mean, standard deviation, and range as a sampling-sensitivity check rather than a definitive confidence interval for the full 2,147-example numeric pool.

\subsection{Repair-Stage Stability Reruns}
\label{app:stability-reruns}

To assess repair-stage variability, we rerun the \gsm repair stage three additional times using the same cached initial traces and the same guarded best-of-3 configuration. The initial traces are fixed, so variation comes from API-side repair generation and formatting/retry behavior. Table~\ref{tab:stability-reruns} reports the original main run together with the three reruns.

\begin{table}[t]
\centering
\scriptsize
\setlength{\tabcolsep}{3pt}
\begin{tabular}{lrrrr}
\toprule
Run & Final Acc. & Fixed & Broken & Accepted \\
\midrule
Main & 96.89 & 17 & 0 & 20 \\
Rerun 1 & 97.04 & 19 & 0 & 22 \\
Rerun 2 & 96.51 & 13 & 1 & 18 \\
Rerun 3 & 96.36 & 12 & 2 & 18 \\
\midrule
Mean & 96.70 & 15.25 & 0.75 & 19.50 \\
Std. & 0.28 & 2.86 & 0.83 & 1.66 \\
\bottomrule
\end{tabular}
\caption{Repair-stage stability on \gsm using fixed cached initial traces. All reruns improve over the 95.60\% initial accuracy, but rare broken-correct cases appear under API-side variation.}
\label{tab:stability-reruns}
\end{table}

\subsection{Acceptance Policy}
\label{app:accept-policy}

The acceptance policy rejects no-op repairs whose final answer does not change. It then considers several acceptance paths, including high-risk semantic repair, empty-generation rescue, very-low-confidence rescue, restricted clean semantic improvement, relaxed support acceptance, and weak-reasoner relaxed acceptance. The strong-reasoner \gsm relaxed path accepts meta-clean or \code{low\_symbolic\_coverage} candidates only when they are graph-clean, sufficiently high-scoring, and equation-supported. The main \gsm configuration does not use the relaxed \code{missing\_constraint} path, because this category may indicate a genuinely omitted quantity. In the ablation, relaxing \code{missing\_constraint} produces the same final accuracy but accepts more modifications without fixing more errors.

\subsection{Thresholds}
\label{app:thresholds}

Table~\ref{tab:appendix-thresholds} summarizes the main thresholds used by the final pipeline.

\begin{table}[t]
\centering
\scriptsize
\begin{tabular}{@{}lr@{}}
\toprule
Parameter & Value \\
\midrule
Best-of-$N$ candidates & 3 \\
Repair trigger graph threshold & 0.80 \\
Graph accept minimum score & 0.60 \\
Graph minimum score drop tolerance & 0.05 \\
Medium trigger meta-score threshold & 0.65 \\
Missing-constraint trigger score & $<0.90$ \\
Minimum repair length & 20 characters \\
Repair max tokens & 768 \\
Format-retry max tokens & 512 \\
Temperature & 0.0 \\
\bottomrule
\end{tabular}
\caption{Main thresholds and decoding settings.}
\label{tab:appendix-thresholds}
\end{table}

\subsection{Ablation Configurations}
\label{app:ablation-configs}

The \gsm ablations are implemented by environment variables. \code{LLM\_REPAIR\_NUM\_CANDIDATES} controls $N$. \code{ENABLE\_GRAPH\_GUARD=false} disables the semantic graph guard. \code{DISABLE\_EQUATION\_SUPPORT\_GUARD=true} disables the equation-support requirement. \code{RELAX\_MISSING\_CONSTRAINT\_ACCEPT=true} enables an ablation-only acceptance path for \code{missing\_constraint} candidates. These ablations are not used in the main configuration unless explicitly stated.

The direct strong-model baselines are implemented separately from the main repair pipeline. \code{solve\_all} re-solves every \gsm example with the strong repair model and accepts all outputs. \code{solve\_triggered} re-solves only examples triggered by the main \method run and accepts all outputs. \code{direct\_bestof3\_gated} uses the same triggered set, strong model, $N=3$ budget, and acceptance gates as \method, but removes the initial trace and diagnostic hint from the repair prompt.

\end{document}